\documentclass[letterpaper, 10 pt, conference]{ieeeconf}  

\IEEEoverridecommandlockouts                              

\title{\LARGE \bf Safety-Critical Centralized Nonlinear MPC for Cooperative Payload Transportation by Two Quadrupedal Robots}

\usepackage{amsmath,amssymb,amsfonts}
\usepackage{cite}
\usepackage{graphicx} 
\usepackage{epsfig} 
\usepackage{times} 
\usepackage{xcolor}
\usepackage{balance}
\usepackage{multicol}
\usepackage{array}
\usepackage{multirow}
\usepackage{booktabs}
\usepackage[colorlinks=true, allcolors=blue]{hyperref}

\newcommand{\Real}{\mathbb{R}}
\newcommand{\col}{\textrm{col}}
\newcommand{\Integer}{\mathbb{Z}_{\geq0}}
\newcommand{\diag}{\textrm{diag}}
\newcommand{\identity}{\mathbb I}
\newcommand{\des}{\textrm{des}}

\newcommand{\terminal}{\textrm{terminal}}
\newcommand{\stage}{\textrm{stage}}

\newcommand{\thr}{\textrm{th}}
\newcommand{\Vset}{\mathcal{V}}
\newcommand{\Eset}{\mathcal{E}}
\newcommand{\Xset}{\mathcal{X}}
\newcommand{\Uset}{\mathcal{U}}
\newcommand{\Oset}{\mathcal{O}}
\newcommand{\Ggraph}{\mathcal{G}}
\newcommand{\Sset}{\mathcal{S}}

\newcommand{\Load}{\textrm{L}}
\newcommand{\SRB}{\textrm{SRB}}
\newcommand{\net}{\textrm{net}}
\newcommand{\glob}{\textrm{global}}
\newcommand{\skews}{\mathbb S}
\newcommand{\SO}{\textrm{SO}(3)}
\newcommand{\so}{\mathfrak{so}(3)}

\usepackage{etoolbox}

\makeatletter
\patchcmd{\@makecaption}{\scshape}{}{}{}
\makeatother

\AtBeginEnvironment{equation}{\vspace{-3pt}}
\AtEndEnvironment{equation}{\vspace{-3pt}}
\AtBeginEnvironment{align}{\vspace{-3pt}}
\AtEndEnvironment{align}{\vspace{-3pt}}
\AtBeginEnvironment{alignat}{\vspace{-3pt}}
\AtEndEnvironment{alignat}{\vspace{-3pt}}

\newtheorem{theorem}{\textbf{Theorem}}

\newtheorem{definition}{\textbf{Definition}}

\newtheorem{remark}{\textbf{Remark}}

\author{Ruturaj S. Sambhus$^{1}$, Yicheng Zeng$^{1}$, Kapi Ketan Mehta$^{1}$, Jeeseop Kim$^{2}$, and Kaveh Akbari Hamed$^{1}$
\thanks{The work of R.S.~Sambhus and K.~Akbari Hamed is partially supported by the NSF under Grant 2423725.}
\thanks{$^{1}$R.S.~Sambhus, Y.~Zeng, K. K. Mehta, and K.~Akbari Hamed (\textit{Corresponding Author}) are with the Department of Mechanical Engineering, Virginia Tech, Blacksburg, VA 24061, USA, {\tt\small \{ruturajsambhus, zyicheng, kmehta4, kavehakbarihamed\}@vt.edu}}
\thanks{$^{2}$J.~Kim is with The University of Texas at El Paso, El Paso, TX 79968, USA, {\tt\small jkim16@utep.edu}}
}

\begin{document}

\maketitle
\thispagestyle{empty}
\pagestyle{empty}


\begin{abstract}
This paper presents a safety-critical centralized nonlinear model predictive control (NMPC) framework for cooperative payload transportation by two quadrupedal robots. The interconnected robot–payload system is modeled as a discrete-time nonlinear differential–algebraic system, capturing the coupled dynamics through holonomic constraints and interaction wrenches. To ensure safety in complex environments, we develop a control barrier function (CBF)-based NMPC formulation that enforces collision avoidance constraints for both the robots and the payload. The proposed approach retains the interaction wrenches as decision variables, resulting in a structured DAE-constrained optimal control problem that enables efficient real-time implementation. The effectiveness of the algorithm is validated through extensive hardware experiments on two Unitree Go2 platforms performing cooperative payload transportation in cluttered environments under mass and inertia uncertainty and external push disturbances.
\end{abstract}


\vspace{-0.3em}
\section{Introduction}
\label{sec:Introduction}

Cooperative payload transportation using legged robots has recently gained significant attention due to its potential in applications such as search and rescue, construction, and logistics in complex and unstructured environments. Compared to single-robot systems, multi-robot cooperative transportation enables handling heavier and larger payloads while providing improved robustness and redundancy. However, the coordination of multiple quadrupedal robots carrying a shared payload introduces strong dynamical coupling through physical interactions, making motion planning and control significantly more challenging.

A key challenge in cooperative payload transportation lies in ensuring safe and dynamically feasible motion in cluttered environments while respecting the coupled dynamics between the robots and the payload. The presence of rigid or semi-rigid connections introduces holonomic constraints that couple the motion of all subsystems, leading to a constrained dynamical system with interaction forces and torques exchanged among agents. Furthermore, real-world deployments must account for model uncertainties, external disturbances, and contact constraints, all of which necessitate robust and safety-critical control strategies.

Model predictive control (MPC) has become a prominent framework for trajectory planning and control of legged robots due to its ability to systematically incorporate system dynamics, constraints, and performance objectives within a finite prediction horizon. Existing MPC-based approaches span both reduced-order and full-order modeling paradigms. Reduced-order template models provide simplified representations of locomotion dynamics that enable real-time trajectory optimization~\cite{Full_Koditschek_Template}. Representative examples include the linear inverted pendulum (LIP) model~\cite{kajita19991LIP} and its variants, such as angular momentum LIP~\cite{ALIP}, spring-loaded inverted pendulum (SLIP)~\cite{SLIP}, vertical SLIP~\cite{vLIP_Sreenath}, and hybrid LIP~\cite{HLIP_Ames}, as well as centroidal dynamics~\cite{orin2013centroidal} and single rigid body (SRB) models~\cite{Kim_Wensing_Convex_MPC_01,Wensing_VBL_HJB,Abhishek_Hae-Won_TRO,Leila_Hamed_RAL,pandala2022robust,Ruturaj_Locomanipulation_RAL}. In contrast, full-order nonlinear MPC methods directly utilize the complete robot dynamics to achieve improved accuracy and dynamic performance, albeit at the expense of increased computational complexity~\cite{Patrick_TRO_Review,crocoddyl,ProxDDP_TRO}.


\begin{figure}
    \centering
    \includegraphics[width=\linewidth]{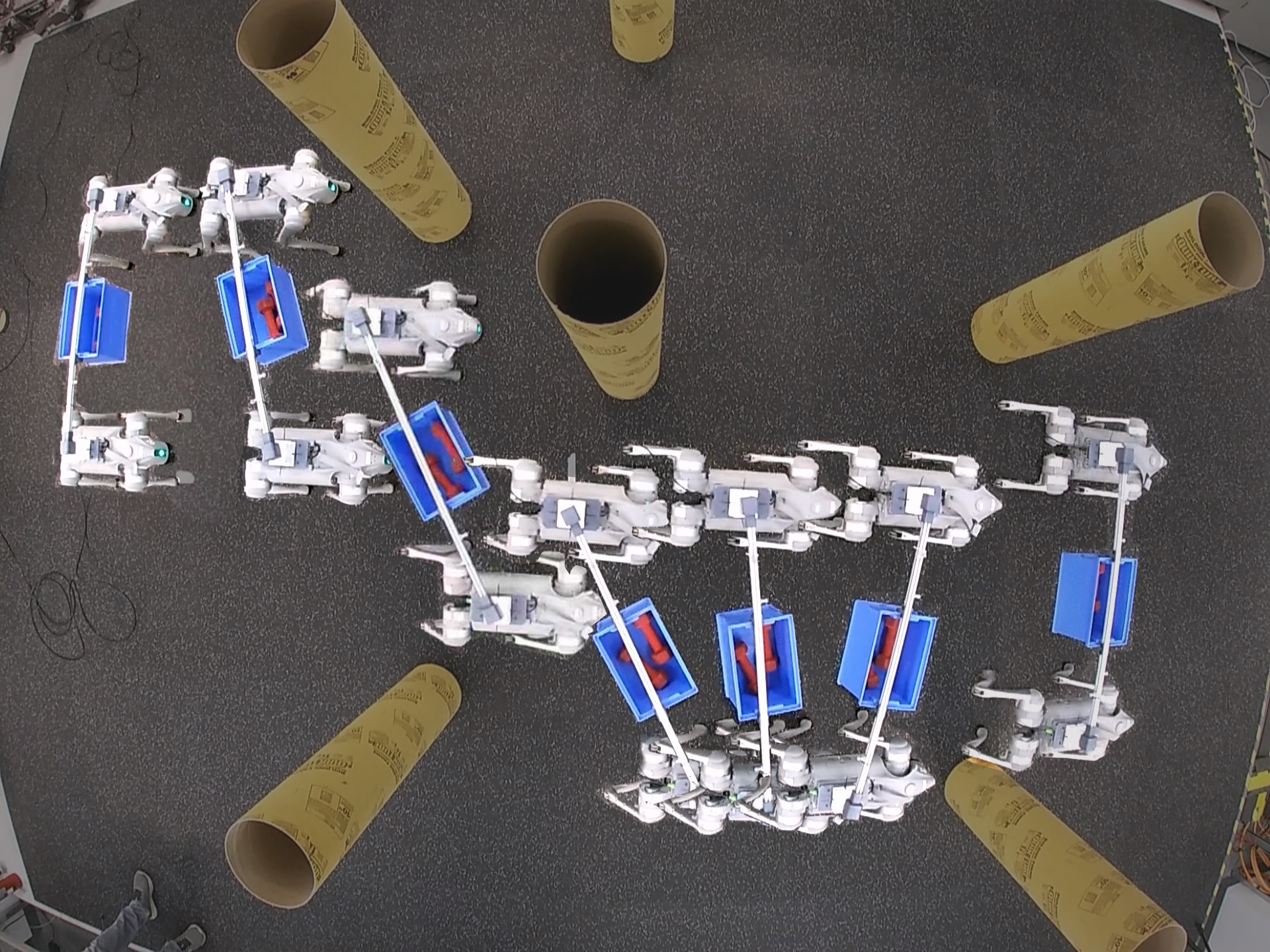}
    \vspace{-1.8em}
    \caption{Snapshot of two Unitree Go2 quadrupedal robots cooperatively transporting a shared payload using the proposed CBF-based centralized NMPC framework in a cluttered environment with cylindrical obstacles. The framework autonomously adjusts the orientations of the robots and the payload to avoid collisions while maintaining safe and stable transportation.}
    \vspace{-2.1em}
    \label{fig:OneSanpshot}
\end{figure}



\begin{figure*}[t]
    \centering
    \includegraphics[width=\linewidth]{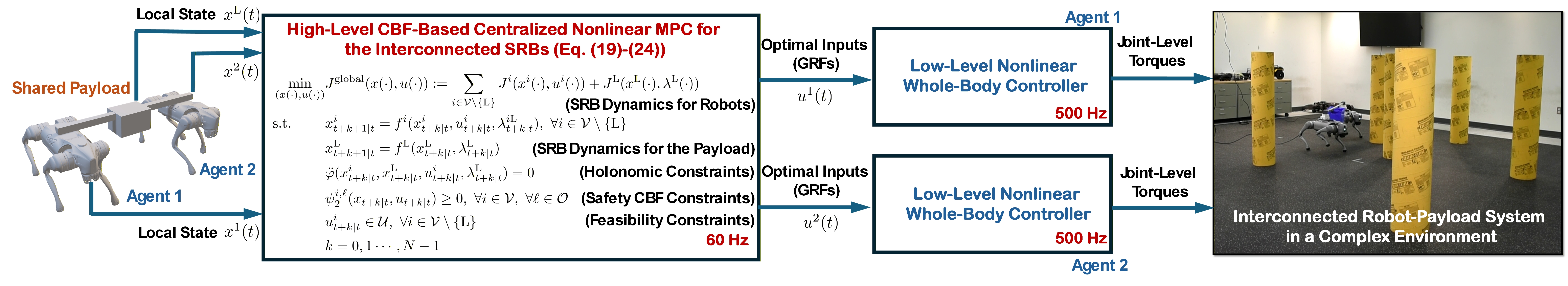}
    \vspace{-1.8em}
    \caption{Overview of the proposed layered control framework. The high-level CBF-based NMPC computes optimal trajectories for the interconnected robot–payload SRB system under holonomic constraints, while low-level nonlinear whole-body controllers enforce full-order robot dynamics.}
    \vspace{-1.7em}
    \label{fig:Overview}
\end{figure*}


Control barrier functions (CBFs) offer a systematic approach to enforcing safety constraints by guaranteeing forward invariance of designated safe sets \cite{CBF_MRS,MRS_CBF_Cavorsi}. Incorporating CBFs within MPC provides a principled framework for safety-critical trajectory optimization by embedding safety constraints directly into the optimal control problem while maintaining feasibility and performance. This paradigm has been successfully demonstrated in both single-robot and multi-robot quadrupedal systems \cite{NMPC_CBF_Sreenath,NMPC_CBF_Ames_Hutter,Basit_RAL,Yicheng_Hamed_IROS}. However, extending these approaches to cooperative payload transportation with shared dynamics remains a significant challenge, as the presence of holonomic constraints introduces strong coupling between robots and the payload, complicating the integration of safety-critical constraints within MPC frameworks.

In this paper, we address this challenge by developing a safety-critical centralized nonlinear model predictive control (NMPC) framework for cooperative payload transportation by two quadrupedal robots. The interconnected robot–payload system is modeled as a discrete-time nonlinear differential–algebraic system that captures both subsystem dynamics and holonomic coupling constraints. Safety is enforced through the integration of higher-order control barrier functions (HOCBFs) within the NMPC formulation, enabling collision avoidance for both the robots and the payload. A key feature of the proposed approach is that the interaction wrenches are retained as decision variables, resulting in a structured DAE-constrained optimal control problem that facilitates efficient real-time implementation. The effectiveness of the proposed framework is validated through extensive hardware experiments, demonstrating robust and safe performance under model uncertainties and external disturbances (see Fig. \ref{fig:OneSanpshot}).


\vspace{-0.3em}
\subsection{Related Work}
\label{sec:Related_work}

Recent works have explored MPC-based approaches for multi-legged robot coordination and cooperative manipulation. In \cite{Jeeseop_TRO,Jeeseop_Hamed_ASME}, a layered control framework was proposed for cooperative locomotion of two holonomically constrained quadrupedal robots using centralized and distributed convex quadratic programming (QP)-based MPC for real-time trajectory planning and whole-body control; however, the dynamics of a shared payload were not explicitly modeled. In our prior work \cite{Randy_ICRA_MultiAgent}, a distributed data-driven predictive control framework was developed for cooperative locomotion of multiple holonomically constrained quadrupedal robots, enabling scalable coordination without relying on explicit model-based optimization; however, payload dynamics and safety-critical constraints were not considered.

The works in \cite{Basit_RAL,Yicheng_Hamed_IROS} develop distributed NMPC frameworks for teams of quadrupedal robots that are not physically interconnected, but instead coupled through inter-agent safety constraints. The work in \cite{JKim_CBF_CooperativeLoco} proposes a layered framework in which CBFs are incorporated at a high-level kinematic planner formulated as a zero-horizon QP to ensure collision avoidance for a team of interconnected quadrupedal robots without explicitly modeling payload dynamics. However, because the CBF constraints are enforced only at the kinematic planning layer, safety is guaranteed only for the reference trajectories provided to the lower-level controllers, and not for the real-time trajectory optimization within the MPC itself. A distributed MPC framework in \cite{Zhou2026ACLM} addresses cooperative loco-manipulation with shared payload dynamics by decomposing a centralized optimal control problem into parallel subproblems via consensus optimization; however, it relies on iterative consensus updates and is primarily validated in simulation. In contrast, the current work develops a centralized safety-critical NMPC framework that explicitly captures holonomic coupling and is validated through real-time hardware experiments.

The work in \cite{DeVincenti2023CLM} formulates a centralized MPC approach for collaborative loco-manipulation using single rigid body dynamics and multiple-shooting trajectory optimization; however, it does not incorporate safety-critical constraints or provide formal safety guarantees, and its validation is limited to simulation. The work in \cite{An2025CLM} proposes a hierarchical reinforcement learning framework with a dynamic reward curriculum for collaborative pick-and-place tasks using legged manipulators, focusing on learning-based coordination rather than model-based control. In contrast, the current work develops a model-based, safety-critical NMPC framework that explicitly captures robot--payload dynamics and enforces safety constraints with real-time guarantees.


\vspace{-0.3em}
\subsection{Contributions}
\label{sec:Contributions}

The main \textit{contributions} of this paper are as follows. We develop a safety-critical centralized NMPC framework for cooperative payload transportation by two quadrupedal robots, modeling the interconnected system as a discrete-time nonlinear differential--algebraic equation (DAE) system with holonomic constraints. We incorporate higher-order control barrier functions (HOCBFs) to enforce collision avoidance for both the robots and the payload, thereby providing formal safety guarantees. We formulate a structured DAE-constrained optimal control problem that retains interaction wrenches as decision variables, enabling efficient real-time implementation (see Fig. \ref{fig:Overview}). The proposed framework is validated through real-time hardware experiments on two Unitree Go2 robots that cooperatively and safely transport a payload, demonstrating robust performance under uncertainties and external disturbances.


\vspace{-0.3em}
\section{Interconnected Dynamical Model}
\label{sec:Interconnected_Dynamical_Model}

\begin{figure}[t]
    \centering
    \includegraphics[width=0.75\linewidth]{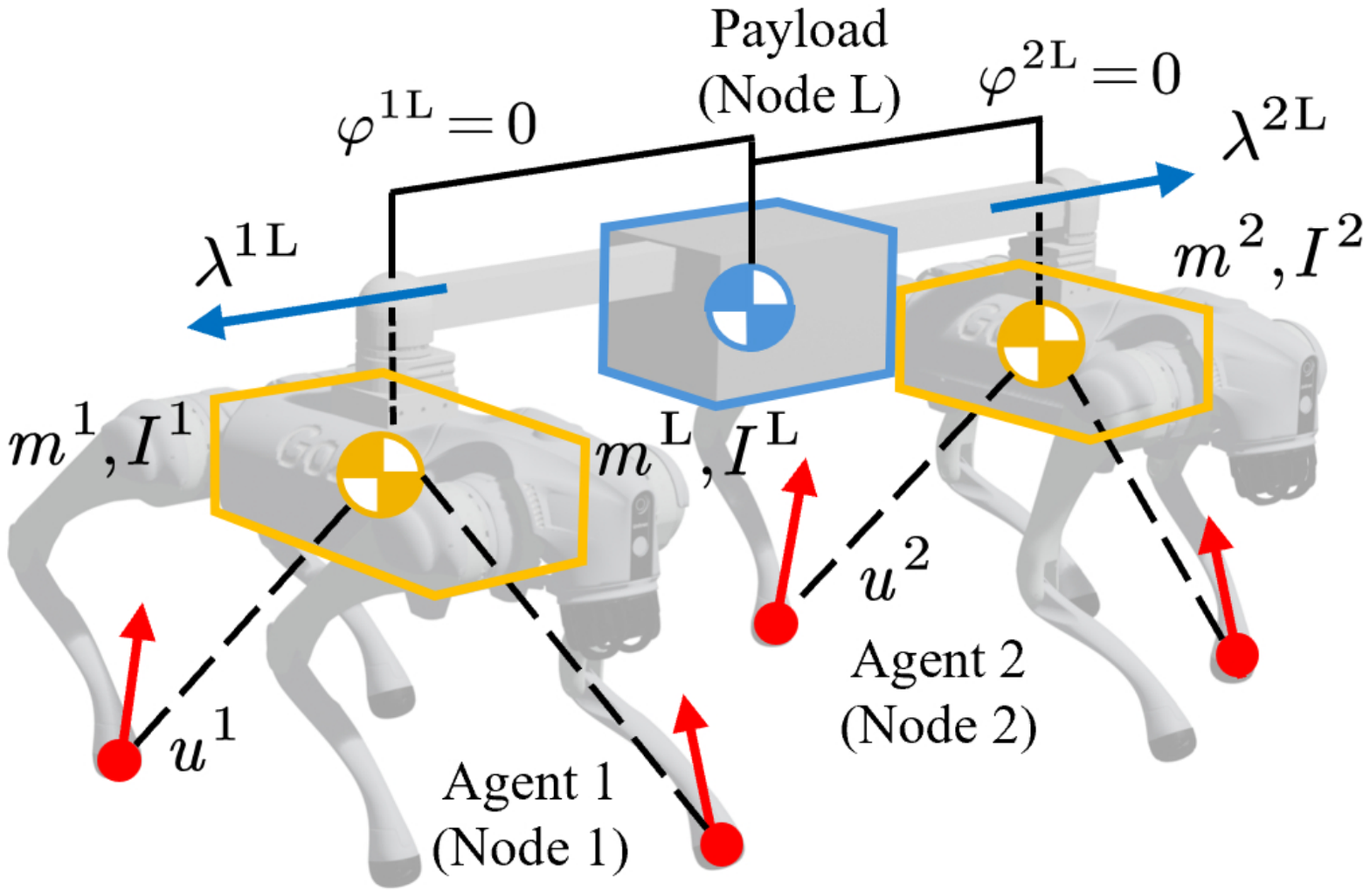}
    \vspace{-1.1em}
    \caption{Illustration of the interconnected SRB model of the robotic agents and the shared payload, showing the rigid holonomic coupling constraints and interaction wrenches.}
    \vspace{-1.7em}
    \label{fig:interconnected_SRB}
\end{figure}

In this section, we present the dynamical model of the interconnected robot--payload system, including the SRB dynamics, interaction wrenches, and holonomic coupling constraints. We consider a system composed of three interconnected single rigid bodies (SRBs) arranged in an open-chain structure. The two side SRBs correspond to the quadrupedal robotic agents, while the middle SRB represents the shared payload (see Fig. \ref{fig:interconnected_SRB}). The interconnection between these bodies is described by an undirected graph $\Ggraph(\Vset,\Eset)$, where the node set is defined as $\Vset := \{1,2,\Load\}$. Nodes $1$ and $2$ correspond to the robotic agents, and node $\Load$ corresponds to the payload. The edge set is defined as $\Eset := \{\{1,\Load\}, \{2,\Load\}\}$, which captures the physical coupling between each robot and the payload. In our notation, quantities associated with node $i \in \Vset$ are denoted using the superscript $i$. Furthermore, quantities associated with an edge $\{i,j\} \in \Eset$ are denoted using the superscript $ij$.


We assume that the state and control input associated with each robotic agent $i \in \Vset \setminus \{\Load\}$ are denoted by $x^{i}(t) \in \Xset \subset \Real^{n_x}$ and $u^{i}(t) \in \Uset \subset \Real^{n_u}$, respectively. For the SRB dynamics, the state vector consists of the Cartesian position of the center of mass (CoM) $p^{i}$, the orientation of the body parameterized by Euler angles $\theta^{i}$, the linear velocity of the CoM $\dot{p}^{i}$, and the angular velocity of the body $\omega^{i}$, all expressed in the world frame. Specifically,
\begin{equation}
    x^{i} := \col(p^{i}, \theta^{i}, \dot{p}^{i}, \omega^{i}),
\end{equation}
where $\col(\cdot)$ denotes the column stacking operator. The control input $u^{i}$ represents the ground reaction forces (GRFs) applied at the stance feet of robot $i$. Similarly, we define the state of the shared payload by $x^{\Load}$. Unlike the robotic agents, the payload does not interact with the ground and does not have actuation through contact forces. Instead, its dynamics are influenced by the interaction wrenches exchanged with the robotic agents, denoted by $\lambda^{i\Load}\in\Real^{n_{\lambda}}$ for $i \in \Vset \setminus \{\Load\}$. The interaction wrenches represent the forces and torques at the connection points and will be further discussed in Remark \ref{remark:interaction_wrenches} (see Fig.~\ref{fig:interconnected_SRB}).

Because the interaction wrenches also act on the robotic agents, the SRB dynamics of each agent can be modeled by the following discrete-time nonlinear system:
\begin{equation}\label{eq:agent_dyn}
    x^{i}(t+1) = f^{i}\left(x^{i}(t), u^{i}(t), \lambda^{i\Load}(t)\right), \quad i \in \Vset \setminus \{\Load\}.
\end{equation}
Here, $t \in \Integer := \{0,1,\cdots\}$ denotes the discrete-time index, and $f^{i}$ is a nonlinear mapping obtained via time discretization (e.g., using the Euler or Runge--Kutta method). Similarly, the payload dynamics are described by the discrete-time nonlinear model
\begin{equation}\label{eq:payload_dyn}
    x^{\Load}(t+1) = f^{\Load}\left(x^{\Load}(t), \lambda^{1\Load}(t), \lambda^{2\Load}(t)\right).
\end{equation}

\textbf{SRB Dynamics:} We describe the continuous-time SRB dynamics used in our formulation for both the robotic agents and the payload as follows:
\begin{equation}\label{eq:SRB_dyn}
    \Sigma^{\SRB}: \begin{cases}
    \ddot{p}^{i}      = \dfrac{f^{i,\net}}{m^{i}} - g_0, \\
    \dot{\theta}^{i}  = A(\theta^{i})\,\omega^{i}, \\
    \dot{\omega}^{i}  = (I^{i})^{-1} \left(\tau^{i,\net} - \skews(\omega^{i}) \, I^{i}\, \omega^{i} \right),
\end{cases}
\end{equation}
where $m^{i}$ denotes the mass of SRB $i$, $g_{0} \in \Real^{3}$ is the gravitational acceleration vector, and $I^{i} \in \Real^{3 \times 3}$ is the inertia matrix of SRB $i$ expressed in the world frame as $I^{i} = R(\theta^{i})\, I^{iB}\, R^\top(\theta^{i})$, with $I^{iB}$ being the inertia in the body frame. Here, $R(\theta^{i}) \in \SO$ denotes the rotation matrix, and $\skews(\cdot): \Real^3 \to \so$ denotes the skew-symmetric operator. The net force and torque applied at the CoM of SRB $i$ are denoted by $(f^{i,\net}, \tau^{i,\net})$.

For robotic agents $i \in \Vset \setminus \{\Load\}$, the net wrench is given by a linear combination of the individual GRFs and the interaction wrench with the payload. Specifically, matrices $E^{i}(t)$ and $F^{i\Load}$ map the GRFs and interaction wrench to the net wrench as
\begin{equation}\label{eq:net_wrench_robot}
    \begin{bmatrix}
        f^{i,\net} \\
        \tau^{i,\net}
    \end{bmatrix}
    =
    E^{i}(t)\,u^{i}(t) + F^{i\Load}\,\lambda^{i\Load}(t).
\end{equation}
We remark that this model is valid if the local GRFs for the robotic agents belong to the friction cone, that is $\Uset=\mathcal{FC}$, where $\mathcal{FC}$ represents the friction cone. For the shared payload, the net wrench is determined solely by the interaction wrenches according to Newton's third law:
\begin{equation}\label{eq:net_wrench_load}
    \begin{bmatrix}
        f^{\Load,\net} \\
        \tau^{\Load,\net}
    \end{bmatrix}
    = - \sum_{i \in \Vset \setminus \{\Load\}} F^{i\Load}\,\lambda^{i\Load}(t).
\end{equation}

\begin{figure}[t]
    \centering
    \includegraphics[width=0.75\linewidth]{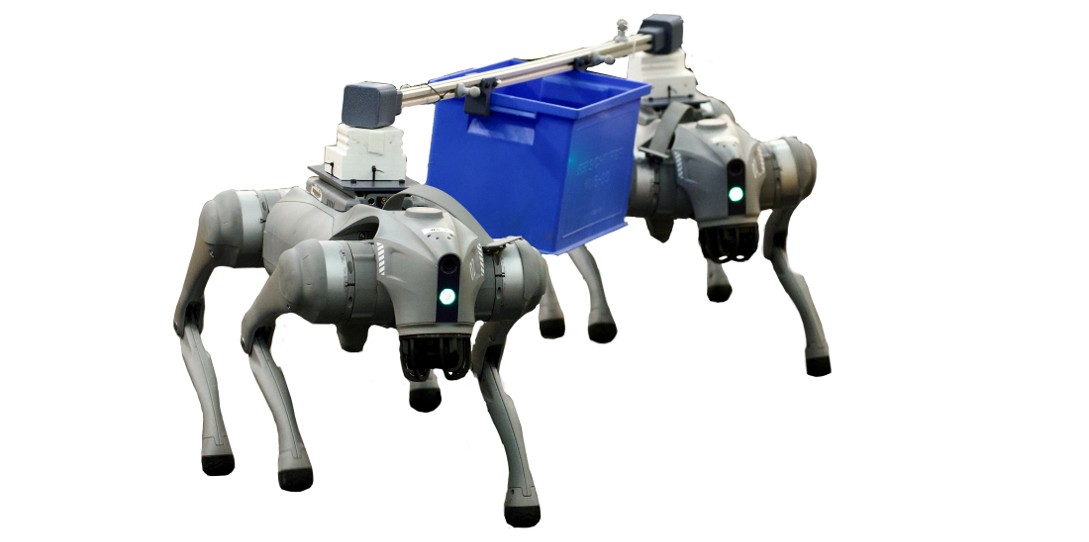}
    \vspace{-1.1em}
    \caption{Illustration of the rigid mechanism between the robots and the payload, with additional masses placed in the basket. The connection enforces translational rigidity, restricts relative roll and pitch motions, and allows relative yaw motion between the robots and the payload.}
    \vspace{-1.7em}
    \label{fig:holonomic_constraints}
\end{figure}

\textbf{Holonomic Constraints:} We consider a rigid connection mechanism between each robot and the payload, as shown in Fig.~\ref{fig:holonomic_constraints}. This mechanism imposes a holonomic constraint between the robotic agents and the payload. In particular, the connection enforces rigidity in translation while restricting the relative roll and pitch motions between the robots and the payload, and allowing free relative yaw motion. 


This connection induces a five-dimensional holonomic constraint between each robotic agent SRB and the SRB of the shared payload, corresponding to three translational constraints and two rotational constraints (roll and pitch), while allowing free relative yaw motion. These holonomic constraints can be expressed compactly as
\begin{equation}\label{eq:holonomic_const}
    \varphi^{i\Load}\left(x^{i}(t), x^{\Load}(t)\right) = 0, 
    \quad \forall i \in \Vset \setminus \{\Load\},
\end{equation}
where $\varphi^{i\Load} : \Xset \times \Xset \to \Real^{5}$ represents the constraint function enforcing (i) coincidence of the interaction points in position and (ii) alignment of the roll and pitch components of the relative orientation between the robot and the payload. Differentiating the holonomic constraints twice along the system dynamics yields
\begin{equation}\label{eq:algebraic_holonomic_const}
    \ddot{\varphi}^{i\Load}\left(x^{i}(t), x^{\Load}(t), u^{i}(t), \lambda^{1\Load}(t), \lambda^{2\Load}(t)\right) = 0, \quad \forall i \in \Vset \setminus \{\Load\}
\end{equation}
which imposes an equality constraint on the control inputs and interaction wrenches that must be satisfied for all $t \geq 0$.

\begin{remark}\label{remark:interaction_wrenches}
The interaction wrenches $\lambda^{i\Load}$ have the same dimension as the corresponding holonomic constraints, i.e., $\dim(\lambda^{i\Load}) = \dim(\varphi^{i\Load}) = 5$ for all $i\in\Vset\setminus\{\Load\}$. More specifically, the interaction wrenches can be obtained by solving the algebraic constraints \eqref{eq:algebraic_holonomic_const}. However, we do not pursue this approach in this paper, as substituting the resulting expressions for the interaction wrenches in terms of the states and local control inputs into the agent and payload dynamics \eqref{eq:agent_dyn} and \eqref{eq:payload_dyn} would increase the nonlinearity of the model and complicate the resulting optimal control problem. Instead, we adopt an alternative approach, presented in Section~\ref{sec:proposed_approach}, in which the interaction wrenches are retained as decision variables in the NMPC formulation. The NMPC then solves for the interaction wrenches while enforcing the algebraic constraints \eqref{eq:algebraic_holonomic_const}.
\end{remark}

\textbf{DAE-constrained Nonlinear Global Model:} For compact notation, we define the global state and input vectors as
\begin{equation}
x := \col\{x^{i}\}_{i\in\Vset}, 
\quad
u := \col\{(u^{i}, \lambda^{i\Load})\}_{i\in\Vset\setminus\{\Load\}}.
\end{equation}
The sets of feasible global states and admissible inputs are denoted by $\Xset^{\glob}$ and $\Uset^{\glob}$, respectively. We then consider the global dynamics of the interconnected robot--payload system obtained by stacking the local dynamics \eqref{eq:agent_dyn} and \eqref{eq:payload_dyn}, together with the algebraic constraints \eqref{eq:algebraic_holonomic_const}. This yields the following compact representation:
\begin{equation}\label{eq:global_dyn}
    \Sigma^{\glob}:\begin{cases}
        x(t+1) = f(x(t),u(t)), \\
        \ddot{\varphi}(x(t),u(t)) = 0,
    \end{cases}
\end{equation}
where $f := \col\{f^{i}\}_{i\in\Vset}$ denotes the global dynamics of the two robotic agents and the payload, and $\ddot{\varphi} := \col\{\ddot{\varphi}^{i\Load}\}_{i\in\Vset\setminus\{\Load\}}$ encodes the holonomic constraints between the robots and the payload. Here, the input $u$ includes both the ground reaction forces and the interaction wrenches. The global model in \eqref{eq:global_dyn} defines a discrete-time nonlinear differential–algebraic system with algebraic coupling constraints, leading to a DAE-constrained nonlinear optimal control problem for trajectory generation.


\vspace{-0.3em}
\section{CBFs for the Interconnected System}
\label{sec:CBFs_for_the_Interconnected_System}

Building on the system model developed in Section \ref{sec:Interconnected_Dynamical_Model}, we develop a safety framework using control barrier functions (CBFs) and higher-order CBFs to systematically enforce collision avoidance for the interconnected robot--payload system. We consider a set of static obstacles indexed by $\Oset := \{1,\cdots,n_{\Oset}\}$, where $o^{\ell} \in \Real^{2}$ denotes the Cartesian position of obstacle $\ell \in \Oset$. The objective is to develop a real-time, safety-critical NMPC scheme that tracks a reference trajectory while respecting the interconnected system dynamics and feasibility constraints, and ensuring collision avoidance for the robot--payload system.

Let $g : \Xset \rightarrow \Real^{2}$ be a continuous mapping such that $g(x^{i})$ gives the projection of the Cartesian coordinates of the CoM of SRB $i$ onto the $xy$-plane. To formally encode safety, we define the \textit{agent--obstacle safe set} as
\begin{equation}
\mathcal{S} 
:= \Big\{ x \in \Xset^{\glob} \ \mid \ 
\|g(x^{i}) - o^{\ell}\| \ge d_{\thr}, 
\ \forall i \in \Vset,\ \forall \ell \in \Oset \Big\},
\end{equation}
where $\|\cdot\|$ denotes the Euclidean norm and $d_{\thr} > 0$ is a prescribed safety margin.

To establish the proposed real-time safety-critical NMPC algorithm, we briefly review the concepts of CBFs and higher-order CBFs for discrete-time systems. 

\begin{definition}[Discrete-Time CBF {\cite{DT-HOCBF}}]
\label{def_CBF}
A function $h : \Xset^{\glob} \rightarrow \Real$ is said to be a discrete-time CBF for the global dynamics \eqref{eq:global_dyn} if there exists a class $\mathcal{K}$ function $\alpha$ satisfying $\alpha(s) < s$ for all $s > 0$ such that 
\begin{equation}\label{eq:CBF_condition} 
\Delta h\left(x(t), u(t)\right) \geq - \alpha\left(h(x(t))\right), \quad \forall x(t) \in \Xset^{\glob}, 
\end{equation} 
where $\Delta h(x(t), u(t)) := h(x(t+1)) - h(x(t))$.
\end{definition}

For systems with relative degree $r>1$, the control input does not appear in the first-order difference of the CBF $h$. In particular, for the interconnected robot--payload system with force and torque inputs, the relative degree is $r=2$. For a general relative degree $r>1$, we construct a sequence of functions as follows: 
\begin{align} \label{eq:hocbf_series}
    \psi_{0}(x(t)) &:= h(x(t)), \nonumber \\
    \psi_{1}(x(t)) &:= \Delta \psi_{0}\left(x(t), u(t)\right) + \alpha_1\left(\psi_{0}(x(t))\right), \nonumber \\
    &\vdots \nonumber \\
    \psi_{r}(x(t),u(t)) &:= \Delta \psi_{r-1}\left(x(t), u(t)\right) + \alpha_r\left(\psi_{r-1}(x(t))\right),
\end{align}
where $\Delta\psi_{j}(x(t),u(t)):=\psi_{j}(x(t+1))-\psi_{j}(x(t))$ for $0\leq j \leq r-1$, and $\alpha_{j}(\cdot)$ are class $\mathcal{K}$ functions satisfying $\alpha_{j}(s)<s$ for all $s>0$ and $j=1,\cdots,r$. This sequence induces the sets
\begin{equation}\label{eq:hocbf_sets}
    \Sset_{j}:=\{x \in \Xset^{\glob} \mid \psi_{j}(x)\geq0\},\quad 0\leq j \leq r-1.
\end{equation}

\begin{definition}[Higher-Order Discrete-Time CBF {\cite{DT-HOCBF}}]\label{def_HOCBF}
The function $h : \Xset^{\glob} \rightarrow \Real$ is a higher-order discrete-time CBF (HOCBF) of relative degree $r$ if there exist functions $\psi_{j}$ for $j \in \{0,1,\hdots,r\}$ defined by \eqref{eq:hocbf_series} and corresponding sets $\mathcal{S}_{j}$ for $j \in \{0,1,\hdots,r-1\}$ defined by \eqref{eq:hocbf_sets} such that 
\begin{equation}\label{eq:hocbf_condition}
    \psi_{r}(x(t),u(t)) \geq 0
\end{equation}
for all $x(t) \in \bigcap_{j=0}^{r-1} \mathcal{S}_{j}$ and for some admissible control input $u(t)\in\Uset^{\glob}$.
\end{definition}

\begin{theorem}[\textit{HOCBF Condition {\cite{DT-HOCBF}}}]
\label{Thm_HOCBF}
Let $h : \Xset^{\glob} \rightarrow \Real$ be a continuous HOCBF of relative degree $r$ defined on $\bigcap_{j=0}^{r-1}\mathcal{S}_{j}$. If there exists a control input $u(t) \in \Uset^{\glob}$ satisfying the HOCBF condition \eqref{eq:hocbf_condition} for all $x(t) \in \bigcap_{j=0}^{r-1}\mathcal{S}_{j}$, then the set $\bigcap_{j=0}^{r-1}\mathcal{S}_{j}$ is forward invariant for the global system \eqref{eq:global_dyn}.
\end{theorem}

To enforce safety for the multi-robot--payload system, we introduce HOCBFs corresponding to agent--obstacle and payload--obstacle collision avoidance. Specifically, for each SRB $i \in \Vset$ (i.e., each robot and the payload) and each obstacle $\ell \in \Oset$, we define the safety function
\begin{equation}\label{eq:HOCBF_def_global_sys}
    h^{i,\ell}(x^{i}) := \|g(x^{i}) - o^{\ell}\| - d_{\thr}.
\end{equation}
Since the relative degree is $r=2$, the corresponding HOCBF condition for each SRB and each obstacle is given by
\begin{equation}\label{eq:HOCBF_conditions_global_sys}
    \psi_{2}^{i,\ell}(x(t),u(t)) \geq 0, \quad \forall i \in \Vset, \ \forall \ell \in \Oset.
\end{equation}


\vspace{-0.3em}
\section{CBF-based Centralized NMPC}
\label{sec:proposed_approach}

We aim to design a computationally efficient, real-time centralized NMPC algorithm that computes feasible and optimal state--input trajectories for the interconnected system, composed of two SRBs corresponding to the robotic agents and one SRB corresponding to the shared payload, while ensuring satisfaction of the HOCBF-based safety constraints \eqref{eq:HOCBF_conditions_global_sys} for obstacle avoidance. The NMPC formulation retains the interaction wrenches as decision variables, rather than eliminating them via the algebraic constraints, and enforces the holonomic constraints explicitly within the optimization problem.

Let us define the predicted global state and input trajectories at time $t$ over a prediction horizon $N$ as $x(\cdot) := \col\{x_{t+1|t}, x_{t+2|t}, \cdots, x_{t+N|t}\}$ and $u(\cdot) := \col\{u_{t|t}, u_{t+1|t}, \cdots, u_{t+N-1|t}\}$, respectively, where $x_{t+k|t}$ and $u_{t+k|t}$ denote the predicted global state and input at time $t+k$, computed at time $t$. Similarly, the local state, control input, and interaction wrench trajectories are defined as $x^{i}(\cdot)$, $u^{i}(\cdot)$, and $\lambda^{i\Load}(\cdot)$ for $i \in \Vset \setminus \{\Load\}$. For compactness, we define the stacked interaction wrench trajectory as
\begin{equation}
    \lambda^{\Load}(\cdot) := \col\{\lambda^{i\Load}(\cdot)\}_{i \in \Vset \setminus \{\Load\}}.
\end{equation}

We propose the following NMPC algorithm for the interconnected system
\begin{alignat}{4}
&\min_{(x(\cdot),u(\cdot))} && J^{\glob}(x(\cdot),u(\cdot)) := \sum_{i\in\Vset\setminus\{\Load\}} J^{i}(x^{i}(\cdot),u^{i}(\cdot)) \nonumber\\
& && \qquad\qquad\qquad\quad\,\,\, + J^{\Load}(x^{\Load}(\cdot),\lambda^{\Load}(\cdot)) \label{eq:NMPC_cost} \\
& \textrm{s.t.} && x^{i}_{t+k+1|t} = f^{i}(x^{i}_{t+k|t},u^{i}_{t+k|t},\lambda^{i\Load}_{t+k|t}),\,\, \forall i\in\Vset\setminus\{\Load\} \label{eq:NMPC_agent_dyn} \\
& && x^{\Load}_{t+k+1|t} = f^{\Load}(x^{\Load}_{t+k|t},\lambda^{\Load}_{t+k|t}) \label{eq:NMPC_payload_dyn} \\
& && \ddot{\varphi}(x^{i}_{t+k|t},x^{\Load}_{t+k|t},u^{i}_{t+k|t},\lambda^{\Load}_{t+k|t})=0 \label{eq:NMPC_algebraic_const} \\
&  && \psi^{i,\ell}_{2}(x_{t+k|t},u_{t+k|t}) \geq 0,\,\, \forall i\in\Vset,\,\,\forall \ell\in\Oset \label{eq:NMPC_HOCBF} \\
& && u^{i}_{t+k|t} \in \Uset,\,\,\forall i\in\Vset\setminus\{\Load\} \label{eq:NMPC_Feasibility}\\
& && k=0,1\cdots,N-1, \nonumber
\end{alignat}
with the initial condition $x_{t|t} = x(t)$, where $x(t)$ denotes the measured global state at time $t$. Here, $J^{i}(x^{i}(\cdot),u^{i}(\cdot))$ for $i\in\Vset\setminus\{\Load\}$ and $J^{\Load}(x^{\Load}(\cdot),\lambda^{\Load}(\cdot))$ denote the local cost functions for the robotic agents and the payload, respectively, and will be specified subsequently. The equality constraints \eqref{eq:NMPC_agent_dyn} and \eqref{eq:NMPC_payload_dyn} represent the local dynamics of the robotic agents and the payload. The algebraic holonomic constraints enforcing the rigid coupling between the SRBs are captured by \eqref{eq:NMPC_algebraic_const}, which also enables the NMPC to solve for the interaction wrenches, as discussed in Remark~\ref{remark:interaction_wrenches}. The HOCBF constraints for both robot--obstacle and payload--obstacle avoidance are imposed via \eqref{eq:NMPC_HOCBF}. Finally, the admissibility of the control inputs (i.e., GRFs) is enforced through \eqref{eq:NMPC_Feasibility}.

The local cost functions for all SRBs in \eqref{eq:NMPC_cost} are defined as
\begin{alignat}{4}
J^{i}(x^{i}(\cdot),u^{i}(\cdot))
& := J^{i}_{\terminal}\!\left(x^{i}_{t+N|t}-x^{i,\des}_{t+N|t}\right) \nonumber\\
& \quad + \sum_{k=0}^{N-1} 
J^{i}_{\stage}\!\left(
x^{i}_{t+k|t}-x^{i,\des}_{t+k|t},
u^{i}_{t+k|t}
\right)
\end{alignat}
which are convex cost functions penalizing the deviation of the local state trajectory $x^{i}(\cdot)$ from a given reference trajectory $x^{i,\des}(\cdot)$ using quadratic terminal and stage costs, defined as
\[
J^{i}_{\terminal}(x^{i}) = \|x^{i}\|^{2}_{P^{i}}, 
\qquad
J^{i}_{\stage}(x^{i},u^{i}) 
:= \|x^{i}\|^{2}_{Q^{i}} + \|u^{i}\|^{2}_{R^{i}},
\]
where $P^{i} \succ 0$, $Q^{i} \succ 0$, and $R^{i} \succ 0$ are positive definite weighting matrices. For the payload subsystem, the control input $u^{i}(\cdot)$ is replaced by the interaction wrench trajectory $\lambda^{\Load}(\cdot)$ in the corresponding cost function.

\textbf{Layered Control Structure:} The proposed framework adopts a hierarchical control architecture (see Fig. \ref{fig:Overview}). The high-level CBF-based centralized NMPC runs at 60 Hz and computes optimal trajectories for the interconnected SRB states, ground reaction forces (GRFs), and interaction wrenches. The numerical details of the NMPC solver are provided in Section \ref{sec:Setup and Controller Synthesis}. At the low level, a nonlinear whole-body controller (WBC) tracks the desired SRB states and GRF references generated for the robotic agents while enforcing full-order robot dynamics. The WBC, adopted from \cite{Randy_Paper_LCSS,Ruturaj_Locomanipulation_RAL}, is implemented as a real-time QP running at 500 Hz, ensuring consistency with full-body dynamics and actuation limits.


\vspace{-0.3em}
\section{Experiments}
\label{sec:Experiments}

This section reports the performance of the proposed layered control framework through hardware experiments.


\begin{figure*}[t]
    \centering
    \includegraphics[width=1.00\linewidth]{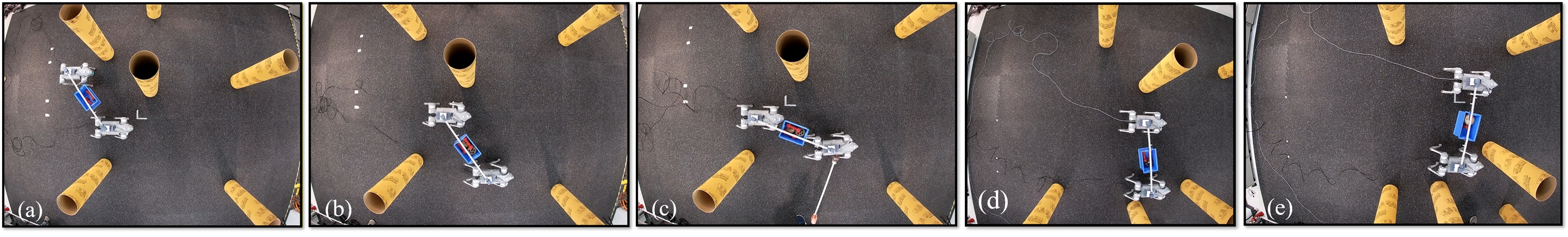}
    \vspace{-2em}
    \caption{Snapshots of all experiments: (a) Experiment~1 with a nominal payload of 5~kg; (b) Experiment~2 with an unmodeled payload of 11.2~kg; (c) Experiment~3 with an unmodeled payload of 11.2~kg under external push disturbances; and (d)–(e) Experiment~4 with different obstacle layouts.}
    \vspace{-1.5em}
    \label{fig:ExpSnapshots}
\end{figure*}


\vspace{-0.3em}
\subsection{Setup and Controller Synthesis}
\label{sec:Setup and Controller Synthesis}

The experimental and simulation studies are conducted using Unitree Go2 quadrupedal robots. Each platform has a mass of approximately 15~kg, a nominal standing height of about 0.28~m, and a total of 18 degrees of freedom (DoFs), comprising 12 actuated joints in the legs (hip roll, hip pitch, and knee pitch per leg) and 6 unactuated floating-base DoFs describing the torso pose. For perception and localization, each robot is equipped with a Unitree L1 4D LiDAR sensor.

The layered control architecture is realized using a multi-threaded implementation on an offboard desktop system equipped with an Intel i9-14900KF CPU and 64~GB of DDR5 RAM, with one thread dedicated to the centralized CBF-based NMPC and two threads assigned to the decentralized low-level WBCs. The robot states are measured in real time using joint encoders and an IMU through the Unitree API, and a kinematic estimator is employed to compute the translational position of the SRB for each robotic agent. 

Due to the rigid bar mechanism, which constrains the relative translation between the two robots and restricts their roll and pitch motions, an alternative kinematic estimator is used to reconstruct the SRB states of the payload from the estimated SRB states of the two robots. Since the yaw motion is not constrained by the rigid connection, the yaw angle of the payload SRB can be uniquely determined from the relative positions of the two robot SRBs once their states are known. Alternatively, a motion capture (MoCap) system can be used to directly and accurately measure the payload SRB states; however, such instrumentation is not employed in this work and is left for future investigation. 

\textit{Real-Time Computation:} The weighting matrices for all SRB subsystem states are selected as $Q^{i}=\diag\{Q^{i}_{p}, Q^{i}_{\theta}, Q^{i}_{\dot p}, Q^{i}_{\omega}\}$, where $Q^{i}_{p} = \diag\{ 1e7, 1e7, 16e7 \}$, $Q^{i}_{\theta} = \diag\{ 1e7, 1e7, 1e7 \}$, $Q^{i}_{\dot p} = \diag\{ 1e6, 1e6, 1e6 \}$, and $Q^{i}_{\omega} = \diag\{ 1e5, 1e5, 1e5 \}$, the terminal cost matrices $P^{i}=10\,Q^{i}$. The control penalty for robotic agents is set as $R^{i}=20\identity$. For the interaction forces, the weighting matrices are defined as $R^{\Load} = \textrm{block }\diag\{R^{\Load}_f, R^{\Load}_\tau\}$, with $R^{\Load}_f = \diag\{ 5e1, 5e1, 5e1 \}$ and $R^{\Load}_\tau = \diag\{ 5e2, 5e2\}$. The prediction horizon is set to $N=8$ with a sampling time of $T_{s}=16.67$ ms. The collision avoidance threshold is chosen as $d_{\thr}=0.5$ m for experiment one and $d_{\thr}=0.6$ m for experiments two and three. The class $\mathcal{K}$ functions for the HOCBF conditions are selected as $\alpha_{1}(s)=0.4s$ and $\alpha_{2}(s)=0.04s$. Trotting gaits are employed for the quadrupedal robots. The CBF-based centralized NMPC is solved at 60 Hz using CasADi \cite{CasADI} with the IPOPT solver \cite{IPOPT}, allowing a maximum of 10 iterations per cycle and warm-starting from the previous solution. The resulting nonlinear program in \eqref{eq:NMPC_cost}--\eqref{eq:NMPC_Feasibility} comprises $596$ decision variables, with the initial SRB states constrained to match the measured state feedback at time $t$. The framework is evaluated in a complex environment with up to $n_{\mathcal{O}}=6$ obstacles using the RaiSim physics engine \cite{RAISIM}. Table~\ref{tab:nmpc_timing} reports the NMPC solve-time statistics (in milliseconds) across three representative experiments. The average solve time of ~11 ms enables real-time implementation at 60 Hz.

\begin{table}[t]
\centering
\caption{Statistics of NMPC solve time (in milliseconds) across three experiments}
\label{tab:nmpc_timing}
\vspace{-1em}
\begin{tabular}{lcc}
\hline
\textbf{Experiment} & \textbf{Mean} & \textbf{Std. Dev.} \\
\hline
Exp.~1 (Nominal Payload) & $11.42$ & $0.6$ \\
Exp.~2 (Robustness to Payload Uncertainty) & $ 11.56 $ & $ 0.42 $ \\
Exp.~3 (Robustness to Disturbances) & $11.45$ & $0.5$ \\
\hline
\end{tabular}
\vspace{-2em}
\end{table}


\vspace{-0.3em}
\subsection{Hardware Experiments}
\label{sec:Hardware Experiments}

We experimentally evaluate the proposed CBF-based centralized NMPC framework on two Unitree Go2 quadrupedal robots connected to a shared payload via the aforementioned rigid mechanism (see Fig. \ref{fig:ExpSnapshots}). The experiments are designed to assess (i) safety-critical cooperative payload transportation in cluttered environments and (ii) robustness to payload mass and inertia uncertainties, as well as external push disturbances. Accompanying experiment videos are available online \cite{YouTube_CBF_Centralized_NMPC}.


\begin{figure}
    \centering
    \includegraphics[width=1.0\linewidth]{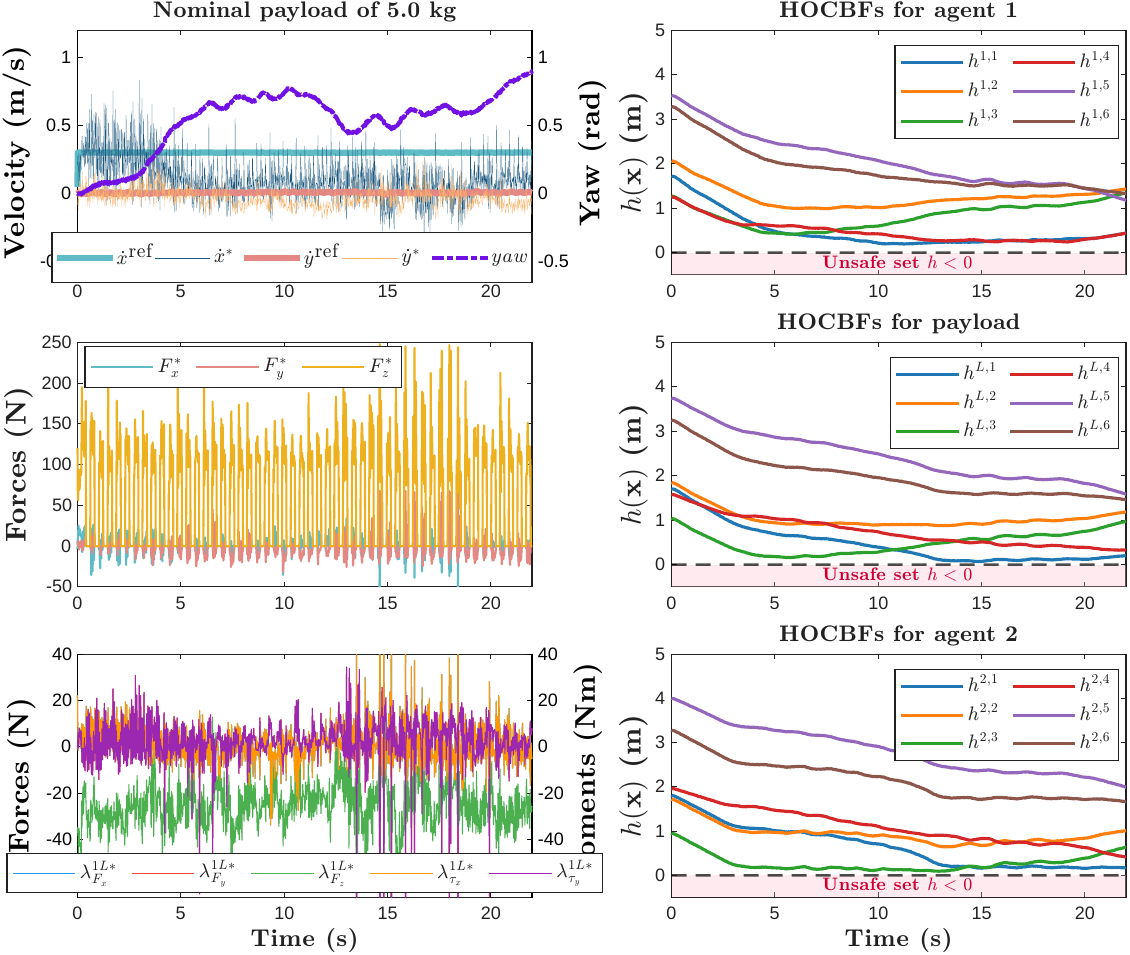}
    \vspace{-1.7em}
    \caption{Reference and actual CoM velocity trajectories along the $xy$-axes for agent~1 during nominal cooperative payload transportation with a 5 kg payload. The yaw plot corresponds to the payload orientation, which is autonomously adjusted by the proposed algorithm to safely navigate through obstacles. The figure also shows the optimal GRFs and interaction wrench profiles (forces and torques) computed by the high-level CBF-based NMPC. The second column illustrates the HOCBF functions $h^{i,\ell}$ for each agent and the six obstacles. All HOCBF functions remain nonnegative throughout the experiment, certifying safety for both the robots and the payload.}
    \vspace{-1.3em}
    \label{fig:Exp1}
\end{figure}


\begin{figure}
    \centering
    \includegraphics[width=1.0\linewidth]{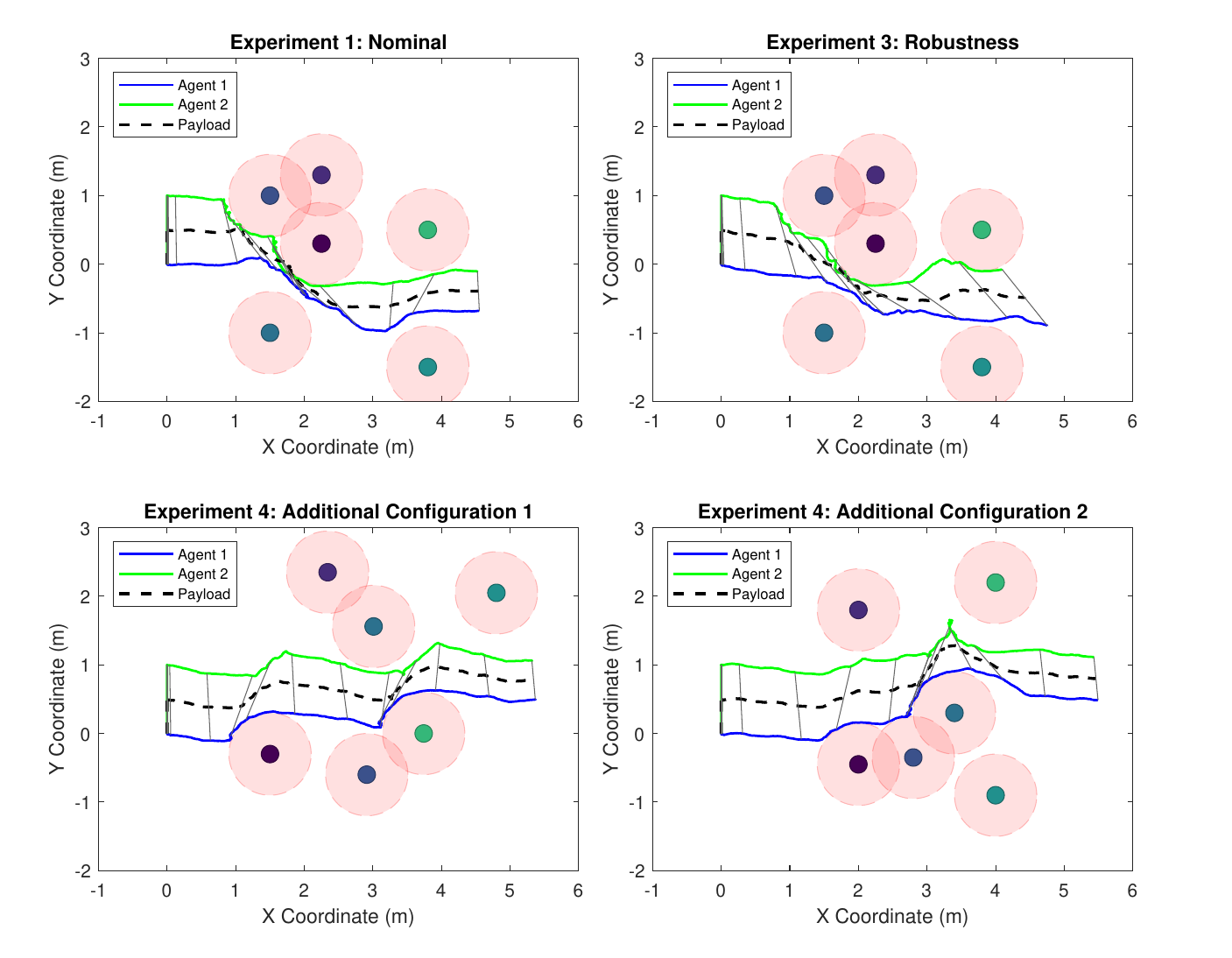}
    \vspace{-1.8em}
    \caption{CoM trajectories of the robotic agents and the shared payload in the $xy$-plane. The rigid mechanism is illustrated by a black line connecting the three systems. The obstacles are represented by solid circles, and the dashed regions around them denote the corresponding safe sets. The plot presents results for Experiments~1,~3, and~4.}
    \vspace{-1.3em}
    \label{fig:Layout}
\end{figure}


\textit{Experiment 1 (Nominal Payload Transportation in Cluttered Environments):} In the first experiment, the payload mass and inertia are set to their nominal values, with a payload mass of 5~kg, which corresponds to approximately 33.33\% of the mass of a single robot (15~kg). The two robots are tasked with navigating a workspace containing six static obstacles while transporting the shared payload (see Fig. \ref{fig:ExpSnapshots}(a)). Individual reference trajectories are assigned to both the robots and the payload with a desired speed of 0.3~m/s. The proposed framework maintains dynamic stability of the interconnected system while tracking the reference trajectories and ensuring collision avoidance for both the robots and the payload. Figure~\ref{fig:Exp1} illustrates the actual CoM velocity trajectories of agent 1 and the yaw trajectory of the shared payload. The algorithm autonomously adjusts the payload orientation to safely navigate through obstacles. The figure also depicts the corresponding optimal GRF and wrench profiles prescribed by the NMPC for the same agent. In addition, it shows the HOCBF functions $h^{i,\ell}$ for all $i \in \Vset$ and $\ell \in \Oset$. The results demonstrate that the interconnected system tracks the prescribed velocity commands whenever it is safe to do so, while all HOCBF functions remain nonnegative, thereby certifying safety throughout the motion. Figure~\ref{fig:Layout} presents the CoM trajectories of the robotic agents and the shared payload in the $xy$-plane.

\textit{Experiment 2 (Robustness to Payload Mass Uncertainty):} In the second experiment, we evaluate the robustness of the proposed framework to payload uncertainty. Specifically, an additional mass is attached to the back of the payload holder, resulting in a total payload mass of 11.2~kg (approximately 74.67\% of the mass of a single robot), while the controller continues to use the nominal payload model of 5~kg from Experiment~1. The robots are tasked with performing the same navigation task in the presence of obstacles (see Fig. \ref{fig:ExpSnapshots}(b)). The results demonstrate that the proposed NMPC framework maintains stable cooperative transportation and continues to track the reference trajectories despite the mismatch between the true and modeled payload dynamics. Furthermore, safety is preserved, as all HOCBF constraints remain satisfied throughout the experiment.  


\begin{figure}
    \centering
    \includegraphics[width=1.0\linewidth]{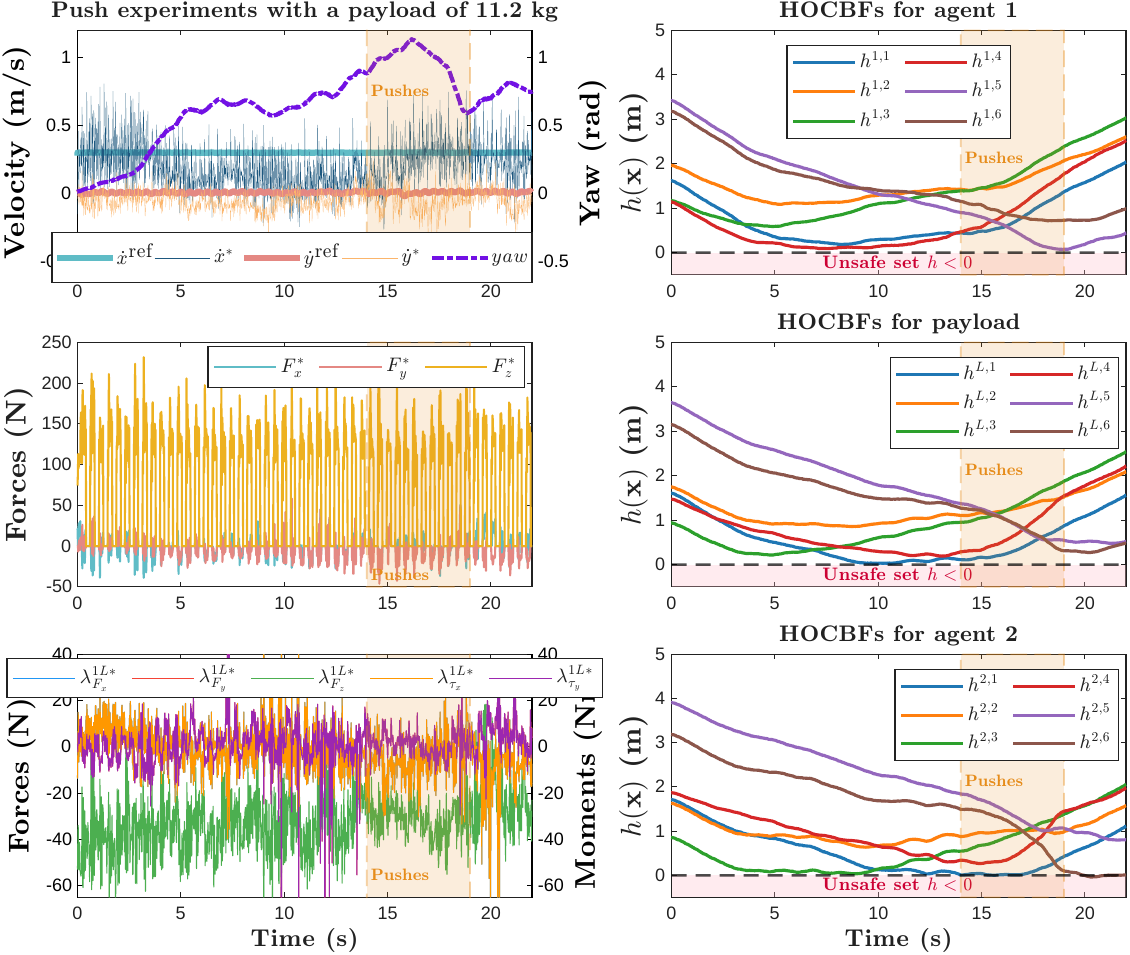}
    \vspace{-1.7em}
    \caption{Reference and actual CoM velocity trajectories along the $xy$-axes for agent~1 during cooperative payload transportation with an uncertain payload of 11.2~kg, subject to unmodeled external push disturbances. The yaw plot corresponds to the payload orientation, which is autonomously adjusted by the proposed algorithm to safely navigate through obstacles. The figure also shows the optimal GRFs and interaction wrench profiles (forces and torques) computed by the high-level CBF-based NMPC. The second column illustrates the HOCBF functions $h^{i,\ell}$ for each agent and the six obstacles, with the time intervals of external pushes highlighted. The HOCBF functions remain nonnegative throughout the experiment, except for a minor and transient violation in $h^{2,6}$ immediately following a push disturbance; the controller subsequently drives agent~2 back to the safe set.}
    \vspace{-1.5em}
    \label{fig:Exp3}
\end{figure}


\textit{Experiment 3 (Robustness to External Disturbances):} The third experiment extends Experiment~2 by introducing external disturbances during cooperative payload transportation. In addition to the unmodeled payload mass, a bystander applies external pushes to one of the robots, resulting in disturbances acting on the interconnected system (see Fig.~\ref{fig:ExpSnapshots}(c)). The results demonstrate that the proposed framework maintains stable cooperative transportation and continues to track the reference trajectories despite the combined effects of payload uncertainty and external disturbances. Furthermore, safety is preserved throughout the experiment, except for a minor and transient violation in $h^{2,6}$ immediately following a push disturbance; the controller subsequently drives agent~2 back to the safe set (see Fig.~\ref{fig:Exp3} and Fig.~\ref{fig:Layout}).

\textit{Experiment 4 (Alternative Obstacle Layouts):} Experiments~1--3 were conducted using a fixed obstacle configuration. To further evaluate the generality of the proposed framework, we perform additional experiments with alternative obstacle layouts (see Fig.~\ref{fig:ExpSnapshots}(d)-(e)). The CoM trajectories of the robots and the payload in the $xy$-plane for these experiments are shown in Fig.~\ref{fig:Layout}.

Overall, the results demonstrate that the proposed framework achieves reliable, safety-critical cooperative transportation under nominal conditions, significant payload uncertainty, and external disturbances, highlighting its robustness and practical applicability. We further observe that the proposed payload-aware NMPC formulation, which explicitly models the payload as a third SRB and incorporates safety constraints via CBFs, enables cooperative transportation of payloads with higher mass relative to the robot body. In particular, compared to our prior work in \cite{Jeeseop_TRO}, which employed a two-SRB model without explicit payload dynamics or safety-critical constraints and relied on a convex MPC formulation, achieving payload transportation up to approximately 55\% of the robot mass, the present framework achieves payload transportation at approximately 74.67\% of the robot mass.


\vspace{-0.3em}
\section{Conclusions}
\label{sec:Conclusions}

This paper presented a safety-critical centralized nonlinear MPC framework for cooperative payload transportation by two quadrupedal robots. The interconnected robot--payload system was modeled as a discrete-time nonlinear DAE system capturing holonomic coupling constraints and interaction wrenches. By integrating higher-order CBFs within the NMPC formulation, the proposed approach ensures collision avoidance for both the robots and the payload while maintaining real-time feasibility. The effectiveness of the framework was demonstrated through hardware experiments on Unitree Go2 platforms operating in cluttered environments under model uncertainties and external disturbances.

Future work will focus on extending the proposed framework to collaborative loco-manipulation tasks by incorporating articulated manipulators on the quadrupedal platforms. In addition, we aim to develop distributed NMPC formulations to improve scalability, along with robust and adaptive control strategies to explicitly account for payload uncertainty.




\bibliographystyle{IEEEtran}
\bibliography{references}

\end{document}